\documentclass[letterpaper, 10 pt, journal, twoside]{IEEEtran}
\ifCLASSINFOpdf
\else
\fi
\usepackage{epsfig} 
\usepackage{times} 
\usepackage{amsmath} 
\usepackage{amssymb}  
\usepackage{graphicx}
\usepackage{xcolor}
\usepackage{siunitx}
\usepackage{array}
\usepackage{diagbox}
\usepackage{adjustbox}
\usepackage{booktabs}
\usepackage{pifont}
\usepackage{amssymb}

\newcommand{\xmark}{\ding{55}}%

\usepackage{url}
\usepackage[
colorlinks=true,
urlcolor=blue,
linkcolor=red,
linkbordercolor=red,
]{hyperref}

\usepackage{subfig}

\usepackage{xcolor}
\definecolor{road}{RGB}{120, 120, 80}
\definecolor{sidewalk}{RGB}{255, 6, 82}
\definecolor{curb}{RGB}{255, 184, 6}
\definecolor{marking}{RGB}{80, 50, 50}
\definecolor{terrain}{RGB}{0, 102, 200}

\newcommand{\etal}{\emph{et~al.~}}

\usepackage{amsmath}

\newcolumntype{P}[1]{>{\centering\arraybackslash}p{#1}}

\DeclareMathAlphabet{\pazocal}{OMS}{zplm}{m}{n}
\newcommand{\unif}{\pazocal{U}}

\begin{document}
%
\title{AutoGraph: Predicting Lane Graphs\\ from Traffic Observations}

\author{Jannik Zürn$^{1}$, Ingmar Posner$^{2}$, and Wolfram Burgard$^{3}$
\thanks{Manuscript received: June 27, 2023; Revised: October 4, 2023; Accepted: November 3, 2023. This paper was recommended for publication by Editor Cesar Cadena Lerma upon evaluation of the Associate Editor and Reviewers’ comments.}
\thanks{This work was supported under DFG grant number BU 865/10-2}
\thanks{$^{1}$Jannik Zürn is with the Faculty of Engineering, University of Freiburg, 79115 Freiburg, Germany
        {\tt\small zuern@cs.uni-freiburg.de}}%
\thanks{$^{2}$Ingmar Posner is with the Applied AI Lab, Oxford University, Oxford, UK, supported by a UKRI/EPSRC Programme Grant [EP/V000748/1]}%
\thanks{$^{3}$Wolfram Burgard is with the Department of Engineering, University of Technology Nuremberg, Nuremberg, Germany}%
\thanks{Digital Object Identifier (DOI): see top of this page.}
}

%
%

\markboth{IEEE Robotics and Automation Letters. Preprint Version. Accepted November 2023}
{Zürn \MakeLowercase{\textit{et al.}}: AutoGraph} 

%



\maketitle

\begin{abstract}
Lane graph estimation is a long-standing problem in the context of autonomous driving. Previous works aimed at solving this problem by relying on large-scale, hand-annotated lane graphs, introducing a data bottleneck for training models to solve this task. To overcome this limitation, we propose to use the motion patterns of traffic participants as lane graph annotations. In our \textit{AutoGraph} approach, we employ a pre-trained object tracker to collect the tracklets of traffic participants such as vehicles and trucks. Based on the location of these tracklets, we predict the successor lane graph from an initial position using overhead RGB images only, not requiring any human supervision. In a subsequent stage, we show how the individual successor predictions can be aggregated into a consistent lane graph. We demonstrate the efficacy of our approach on the UrbanLaneGraph dataset and perform extensive quantitative and qualitative evaluations, indicating that AutoGraph is on par with models trained on hand-annotated graph data. Model and dataset will be made available at \url{http://autograph.cs.uni-freiburg.de/}.
\end{abstract}

\begin{IEEEkeywords}
Deep Learning for Visual Perception, Computer Vision for Transportation, Semantic Scene
Understanding
\end{IEEEkeywords}

%
\IEEEpeerreviewmaketitle


\section{Introduction}

Autonomous vehicles require detailed knowledge about their surroundings to safely and robustly navigate complex environments. Most approaches to automated driving follow one of the two major paradigms: \textit{map-based} or \textit{mapless} driving. Map-based approaches typically rely on HD maps entailing detailed geospatial information relevant to driving tasks, including the positions of traffic lights, lanes, or street crossings. In this context, the graph of lane centerlines (i.e. the lane graph) is a crucial component that encodes the position and connectivity of all lanes. A major bottleneck in deploying map-based autonomous driving approaches is the slow and expensive manual annotation process to generate HD maps for all regions where the vehicle is intended to operate. Methods capable of estimating the lane graphs robustly in an automated fashion are crucial for scaling up the areas covered by HD maps~\cite{homayounfar2019dagmapper, zurn2021lane, buchner2023learning}. Mapless driving approaches, in contrast, solely rely on onboard sensor measurements to infer the position and layout of objects and surfaces relevant to the driving task, including the position and orientation of roads and lanes. For mapless driving, the accurate and robust estimation of the spatial and topological lane layout in the vicinity of the vehicle is paramount for safe and efficient navigation. Therefore, automatic lane graph estimation is a crucial task in map-based and mapless automated driving.

Prior work in lane graph estimation focuses on training models under full supervision~\cite{buchner2023learning, zurn2021lane, he2022lane}, relying on large-scale ground-truth lane graph annotations, typically obtained from a large number of human annotators. The production of accurate annotations such as those available as part of the Argoverse2~\cite{wilson2023argoverse} and NuScenes~\cite{caesar2020nuscenes} datasets is, therefore, resource-intensive in both money and time.

\begin{figure}
    \centering
    \includegraphics[width=0.48\textwidth]{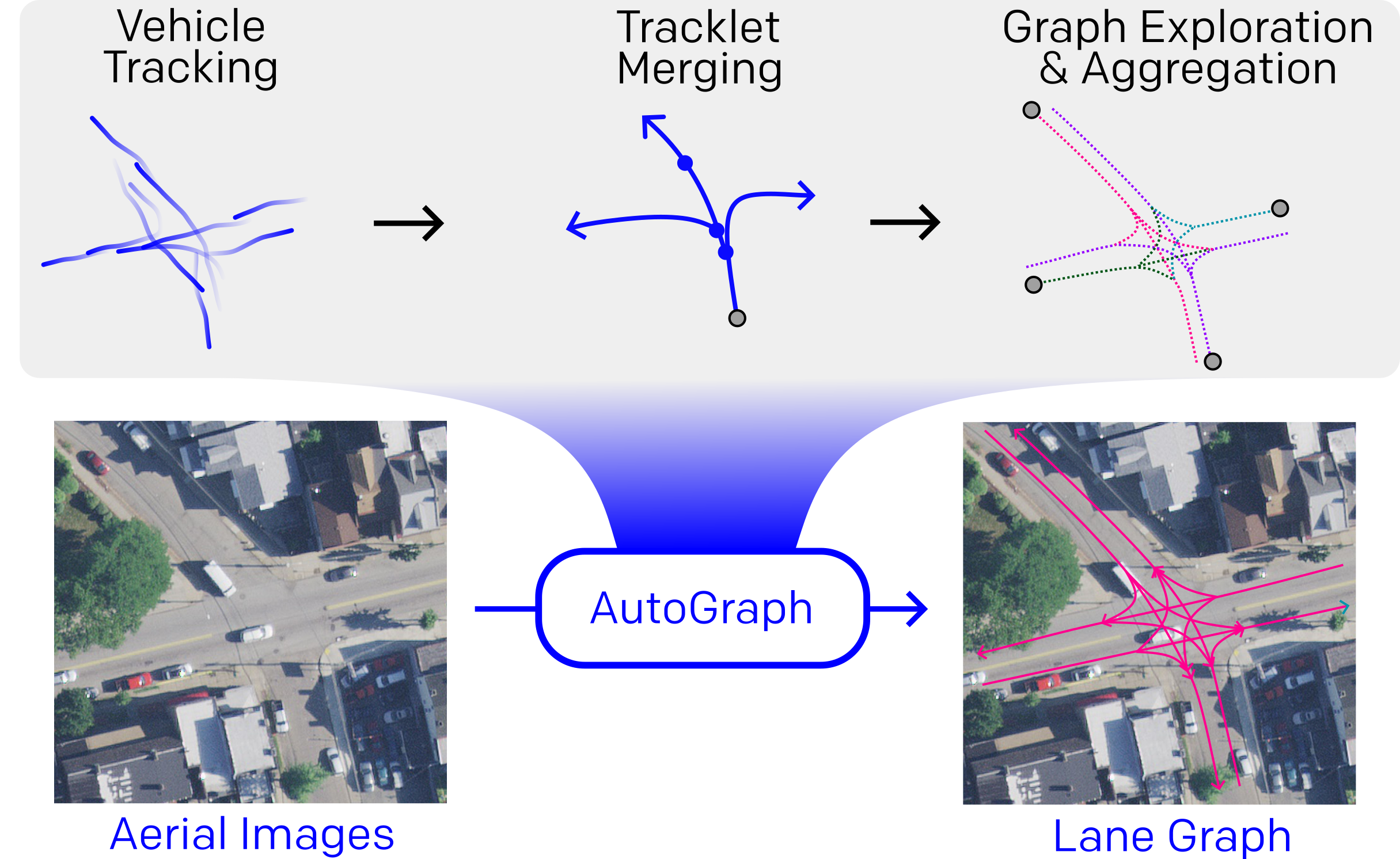}
    \caption{Our approach AutoGraph leverages vehicle tracklets and predicts complex lane graphs from aerial images without requiring any hand-annotated lane graphs for supervision.}
    \label{fig:covergirl}
\end{figure}

Inspired by the success of vehicle tracking approaches~\cite{ondruska2016deep, dequaire2018deep, openpcdet2020} and by prior work in the context of automatic annotation from traffic participants~\cite{barnes2017find, zurn2022trackletmapper}, in this work, we propose to leverage traffic participant tracklets as the only annotation source for lane graph estimation and do not require any manually obtained graph annotations. Most traffic participants follow their respective lanes with high accuracy. Aggregated over large numbers, the trajectories of traffic participants encode the overall structure of lane graphs well (see Fig.~\ref{fig:covergirl}). We interpret this driving data as a data source for the annotation of lane graphs. In our approach, AutoGraph, we track traffic participants in challenging urban environments and propose a novel tracklet merging scheme, allowing us to formulate a supervised learning task in which we leverage aerial images as input and the merged tracklets serve as the learning target for our model. The overall approach is capable of accurately predicting lane graphs covering large areas with high accuracy while the pipeline does not require any hand-annotated data.

To summarize, this work makes the following contributions:
\begin{itemize}
    \item a novel tracklet aggregation scheme leveraging observed traffic participant tracklets as annotation sources for lane graph estimation models;
    \item the large-scale \textit{UrbanTracklet} dataset with hundreds of thousands of vehicle and pedestrian tracklets generated from the Argoverse2 and UrbanLaneGraph datasets;
    \item and extensive qualitative and quantitative ablation studies on the UrbanLaneGraph dataset, demonstrating the efficacy of our approach.
\end{itemize}

\section{Related Works}

\paragraph{Lane Graph Estimation}

Over the past years, the task of lane graph estimation has gained much attention in the autonomous driving research community. In contrast to road graph estimation, where the goal is to estimate the connectivity between roads~\cite{bastani2018roadtracer, tan2020vecroad}, lane graph estimation entails predicting the position of lanes and how they are connected. This task is much more challenging, in particular in areas with complex lane connections such as roundabouts or multi-arm intersections. Homayounfar \etal~\cite{homayounfar2019dagmapper} predict the lane graph of highway scenes with an iterative RNN model from projected LiDAR data. Zürn \etal~\cite{zurn2021lane} proposed a Graph R-CNN-based model for lane graph estimation from aggregated LiDAR data in urban scenes. He \etal~\cite{he2022lane} leverage a multi-stage approach for lane graph extraction in which they first extract straight road sections between intersections and subsequently learn the connectivity between each incoming and outgoing lane arm. Büchner \etal~\cite{buchner2023learning} proposed a bottom-up approach for lane graph estimation. They first estimate the successor graph from a given starting position using a graph neural network and subsequently aggregate a full lane graph by iteratively merging each successor graph into a global one. Similar to our work in spirit, Karlsson \etal~\cite{karlsson2023learning} infer maximum likelihood lane graphs from traffic observations with a directional soft lane probability model. They evaluate their model on the NuScenes dataset. However, they do not consider model inference from aerial images but from aggregated onboard sensor measurements. Crucially, and in contrast to our approach, their model is not capable of estimating large lane graphs due to the non-iterative nature of their approach. 

While the aforementioned works show promising results in challenging environments, most of them require large-scale handcrafted graph annotations or cannot generate predictions for large-scale scenes. In the approach presented here, we do not require any manual annotations and instead leverage data encoded in the behavior of observed traffic participants.

\paragraph{Trajectory Prediction}

Our successor lane graph prediction module is related to the task of trajectory prediction. From the large body of literature available in this field we briefly review the most relevant recent related works. Most approaches condition their models on rasterized or vectorized HD map representations~\cite{chai2019multipath, zhao2021tnt, gao2020vectornet}, discrete graphs~\cite{salzmann2020trajectron++} or aerial images~\cite{zhao2019multi}. In Chai \etal~\cite{chai2019multipath}, future vehicle positions are encoded by estimating the distribution over future trajectories for each agent while Zhao \etal~\cite{zhao2021tnt} leverage a three-stage approach that finds prediction targets, estimates future motion for each, and scores each predicted trajectory to yield the final motion prediction. Our work also shares similarities with the line of work by Gilles~\etal~\cite{gilles2021home, gilles2021thomas, gilles2022gohome}. They frame the trajectory prediction task as a heatmap regression task, where an HD map representation is used for prediction conditioning. After subsequent post-processing of this heatmap, they sample future agent trajectories. 
In contrast to most existing works, we refrain from leveraging an HD map representation and instead solely rely on aerial images for our prediction task. In addition to regressing future possible agent positions, we also use this prediction block as input for a graph aggregation module to learn a complete lane graph of a given input image.

\paragraph{Automatic Annotations in Autonomous Driving}

There exists a sizable body of work that considers the data encoded either in the driving behavior of the ego-vehicle or of other traffic participants. Barnes \etal~\cite{barnes2017find} use the ego-trajectory of a vehicle to annotate drivable regions in an image. They project their own future positions into the current camera image to label pixels as drivable. Other works in self-supervised learning for navigation~\cite{wellhausen2019should, zurn2020self} also use the ego-trajectory to label pixels for a vision-based ground classifier. Tracklets have been used by multiple previous works in the context of autonomous driving tasks. Zürn \etal~\cite{zurn2022trackletmapper} used the trajectories of other traffic participants such as vehicles and pedestrians, obtained from a LiDAR tracker, to annotate ground surfaces in urban environments. Other works also explored the benefits of inferring driving policies from the behavior of other traffic participants~\cite{xu2017end, zhang2021learning, chen2022learning}. Chen \etal~\cite{chen2022learning} leverage driving experiences collected from the ego vehicle and other vehicles jointly to train a driving policy from real-world data. Recent work by Collin \etal~\cite{colling2022hd} proposes an automated system for aggregating observed traffic participant tracklets into a lane graph representation. While their work shows a good performance in dense traffic scenarios, it does not generalize to unseen areas since their approach does not involve training a model on this data.
\section{Technical Approach}

Our approach proceeds in three steps. In the first step, denoted \textit{tracklet parsing and merging}, we track traffic participants through all scenes in the dataset and prepare the data for model training. In the subsequent \textit{model training} step, we train the proposed model with data obtained in the first stage. In the third step, we perform inference with our trained model to perform \textit{graph exploration and aggregation} into a globally consistent representation. In the following, we describe each step in detail.

\begin{figure}
    \centering
    \includegraphics[width=0.48\textwidth, ]{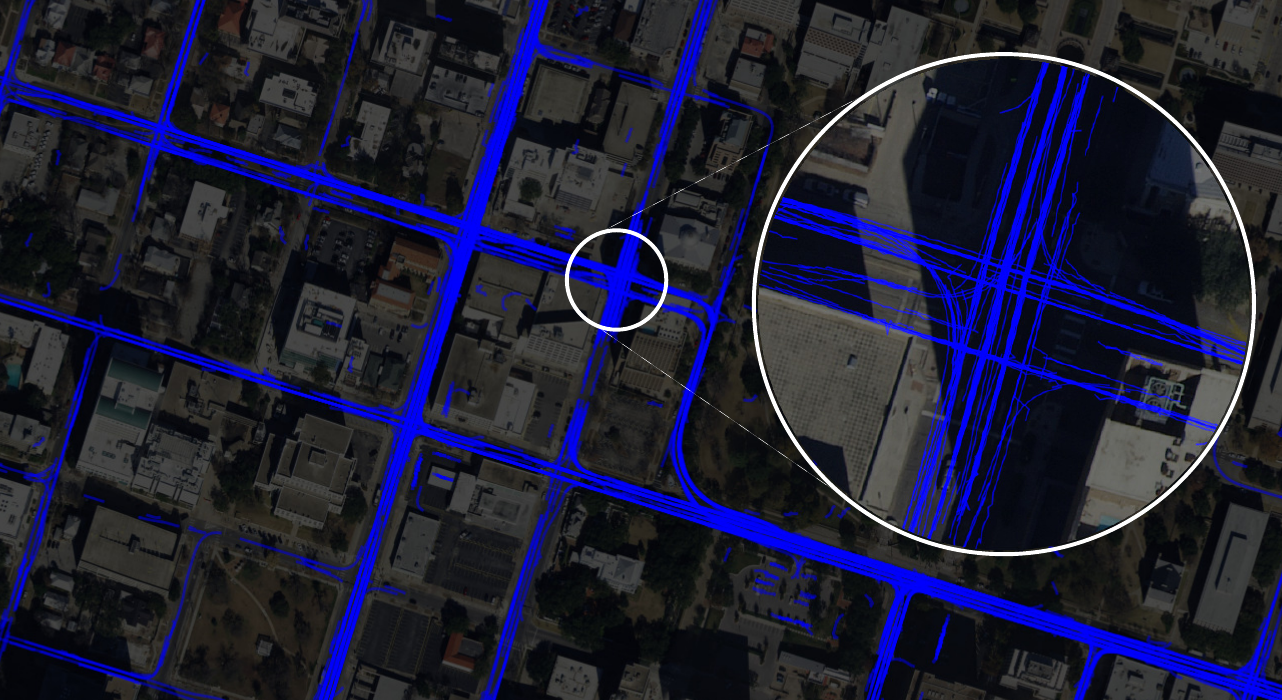}
    \caption{Visualization of tracklets in the city of Austin, Texas, aligned with aerial imagery (darkened for better contrast).}
    \label{fig:tracklets}
\end{figure}

\subsection{Tracklet Parsing and Merging}
\label{subsec:tracklet-parsing}

In the following, we describe our tracklet parsing and merging pipeline. We start our data processing pipeline by tracking traffic participants from ego-vehicle data in all available scenes of the Argoverse2 dataset~\cite{wilson2023argoverse} across all six available cities. Each scene in the dataset consists of approximately 20 seconds of driving. For each scene, we track vehicles such as cars, trucks, motorcycles, and busses using a pre-trained LiDAR-based object detector~\cite{openpcdet2020} that processes the vehicle onboard LiDAR point clouds. We note that the performance of the tracker predictions may affect the quality of the annotations and thus the downstream performance of our approach. We transform all tracklets into a global coordinate frame. Subsequently, we smooth the tracklets with a moving average filter to minimize the amount of observation noise and the influence of erratic driving behavior (i.e., steering inaccuracies). 

Fig.~\ref{fig:tracklets} illustrates an exemplary urban scene with the observed vehicle tracklets. Due to imperfect vehicle driving manoeuvres and the inherent observation noise, tracklets of observed traffic participants do not perfectly overlap with the ground-truth lane graph. Furthermore, and more importantly, each tracklet only covers a subset of the actual lane graph since the corresponding vehicle was only visible from the ego vehicle for a few seconds. Thus, the goal of the tracklet merging module is to merge tracklets that have significant overlap and follow the same underlying lane segment with a high likelihood. 

In the following, we describe the tracklet merging procedure. We define a tracklet $\mathbf{T}$ as a list of points $\mathbf{p}_t \in \mathbb{R}^2$ describing the 2D position of a tracked object centroid at the tracking time step $t$ and a list of heading vectors $\mathbf{h}_t \in \mathbb{R}^2$ describing the heading of the tracked object. The tracking frequency is constant for all tracklets and equals the LiDAR frequency. We define the set of all object positions in all tracklets as $\mathcal{P}$. Our goal is to merge multiple tracklets into a successor graph that encompasses all regions that have been traversed by tracked vehicles which passed through a given starting position. To this end, we define an Euclidean distance tracklet merging matrix $\mathbf{M}_D$ and an angle-based merging matrix $\mathbf{M}_A$. The Euclidean distance merging matrix $\mathbf{M}_D$ is defined as the element-wise Euclidean distance of two object centroids:

\begin{equation}
    M_{D,ij} = || \mathbf{p}_i - \mathbf{p}_j ||_2^2,
\end{equation}
where $\mathbf{M}_D \in \mathbb{R}^{|\mathcal{P}| \times |\mathcal{P}|}$. We also define an angle matrix $\mathbf{M}_A \in \mathbb{R}^{|\mathcal{P}| \times |\mathcal{P}|}$, indicating the absolute relative angle between object heading $\mathbf{h}_i$ and $\mathbf{h}_j$:

\begin{equation}
    M_{A,ij} = \big| \arccos{\Big( \frac{\mathbf{h}_i \cdot \mathbf{h}_j}{|\mathbf{h}_i| \cdot |\mathbf{h}_j|}  \Big)} \big|.
\end{equation}

To merge multiple tracklets into a successor graph, we define a binary tracklet merging matrix $\mathbf{M} \in \{0, 1\}^{|\mathcal{P}| \times |\mathcal{P}|}$ as follows:

\begin{equation}
    \mathbf{M} := [\mathbf{M}_D < d_{max}] \land [\mathbf{M}_A < \alpha_{max}],
\end{equation}

where $M_{ij} = 1$ implies a merging of the tracklet containing $\mathbf{p}_i$ with the tracklet containing $\mathbf{p}_j$. So far, we only considered the relative heading angle and relative position of two tracked objects. In the final step, we update $\mathbf{M}$ and integrate all recorded tracklets into the connectivity matrix. Since each single tracklet consists of a list of observed object positions and headings, we add all pairs of consecutive object positions and headings in each tracklet to $\mathbf{M}$ as well. Note that a connection $p_i \rightarrow p_{i+1}$ encoded in a specific tracklet adds the respective connection in $\mathbf{M}[i, i+1]$ but does not add the connection in $\mathbf{M}[i+1,i]$ since each tracklet has a notion of direction. Thus, $\mathbf{M} \neq \mathbf{M}^T$ in the general case.

\begin{figure}
    \centering
    \includegraphics[trim={0 0 0 3cm},clip,width=0.48\textwidth]{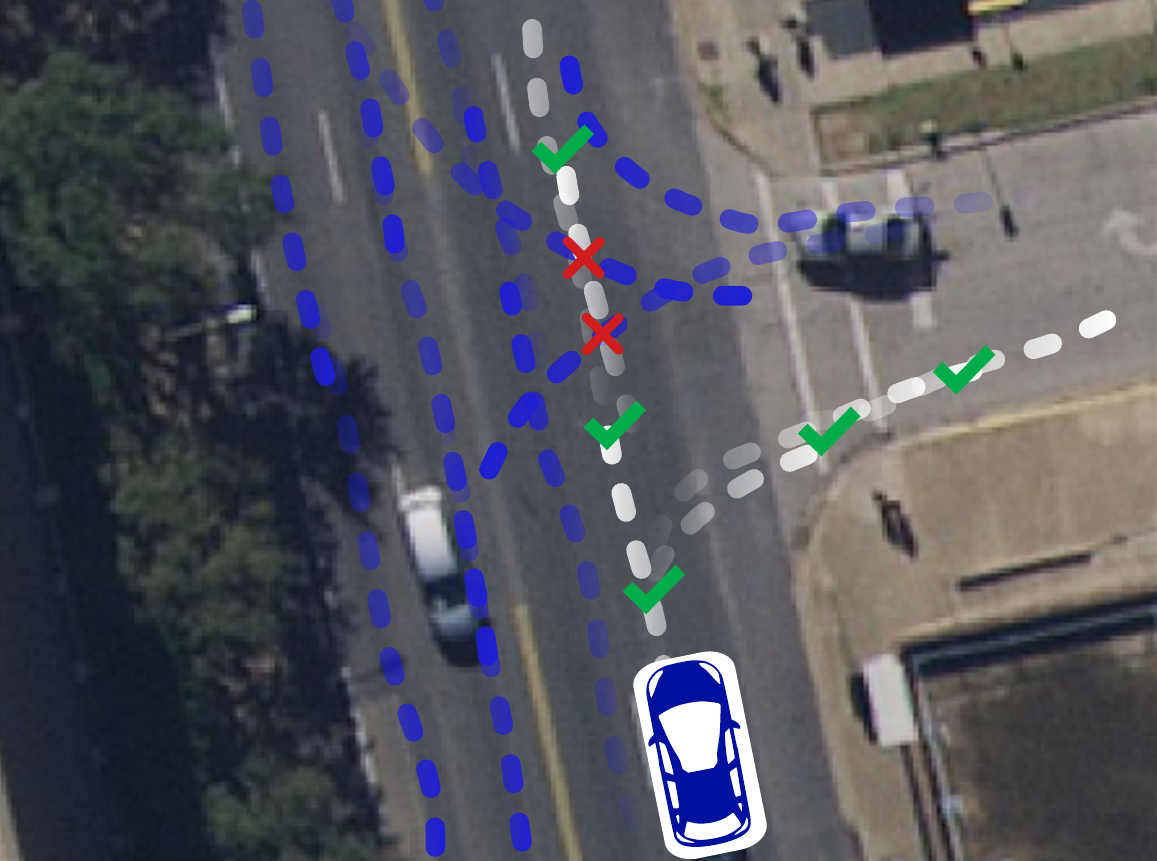}
    \caption{A T-junction with vehicle tracklets (blue dashed lines). Merging points for tracklets are indicated with a green checkmark while exemplary failed merges are indicated with a red cross. The extracted successor graph is visualized in white.}
    \label{fig:trajectory-merging}
\end{figure}

Using this formulation, we now have a mechanism for generating a successor graph $\mathcal{S}_q$ from a query point $\mathbf{q}$ by following all tracklets connected to $\mathbf{q}$ according to $\mathbf{M}$. In order to fit our model to this data, we randomly select a query point $\mathbf{q}$ from the aerial image and extract a small image crop around $\mathbf{q}$. We extract all tracklets visible in this crop and extract the successor graph $\mathcal{S}_q$. For an exemplary visualization of the merging procedure, please refer to Fig.~\ref{fig:trajectory-merging}.

\begin{figure*}
    \centering
    \includegraphics[width=0.94\textwidth]{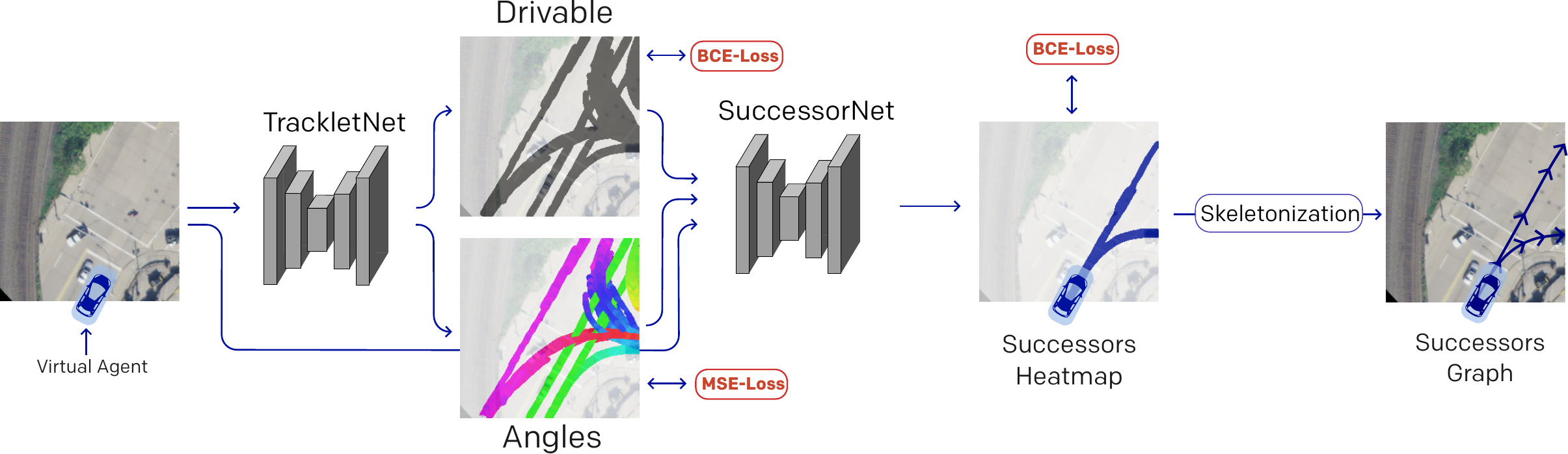}
    \caption{Illustration of our successor graph prediction approach. We first place a virtual agent on a crop of an aerial image. We first predict the \textit{Drivable} and \textit{Angles} map layers from the aerial image crop with a fully convolutional neural network. We subsequently predict the successor lane heatmap from the aerial image crop, the predicted drivable surface, and lane angles. The successor lane heatmap is post-processed into a successor graph, encoding the location of successor lanes and lane split points that are reachable from the current agent's position.}
    \label{fig:approach1}
\end{figure*}

Furthermore, we extract the \textit{Drivable} map layer and the \textit{Angles} map layer. In these layers, we collect all tracklets of the whole city and colorize all pixels that are covered by a tracklet as 1 for the \textit{Drivable} map layer or as the tracklet angle $\alpha$, for the \textit{Angles} layer. The remaining pixels are assigned a value of 0. For visualization of these map layers, please refer to Fig.~\ref{fig:approach1}.

\subsection{Model Training}
\label{subsec:model-training}

After our aggregation step, we are able to query all tracklets that are visible in an aerial image crop, starting from a given querying position $\mathbf{q}$. To obtain a training dataset for our models, for each query pose $\mathbf{q}$, we extract an aerial image crop $RGB_\mathbf{q}$ from the full aerial image, centered and oriented around the query pose. In the same way, we crop the drivable map, producing $D_\mathbf{q}$, and the angle map, producing $A_\mathbf{q}$. The whole training pipeline is visualized in Fig.~\ref{fig:approach1}. Our model consists of two sub-networks. We train a DeepLabv3+ model~\cite{chen2018encoder} to predict the pixel-wise drivable and angle maps from an RGB aerial image input, using $D_\mathbf{q}$ and $A_\mathbf{q}$ as the learning targets. We denote this model as TrackletNet. This initial task is identified as an auxiliary task, leveraging the vast amount of tracklets readily available for a given crop. For training, we use a binary cross-entropy loss to guide the prediction of the drivable map layer and a mean squared error loss for the prediction of the angle map. We encode the \textit{Drivable} layer as a tensor $D_{ij} \in \{0,1\}^{H \times W}$. To circumvent the discontinuous angles at the singularity $\alpha = \pm \pi$, we encode the angle at pixel location $(i,j)$ as a value pair $[\sin(\alpha), \cos(\alpha)]^T$, producing the \textit{Angles} layer $A^k_{ij} \in \mathbb{R}^{H \times W \times 2}$. To summarize, during the TrackletNet training stage, we minimize the following loss:

\begin{equation}
    \mathcal{L} = \frac{1}{HW} \sum_{i<H} \sum_{j<W} \Big( D_{ij}  \log \hat{D}_{ij} + \alpha \sum_{k} || A^k_{ij} - \hat{A}^k_{ij}||_2^2  \Big),
\end{equation}

 with a weighing factor $\alpha$ between the drivable surface classification and the angle regression. In our experiments, we set $\alpha = 1$.
 
Subsequently, we train a separate DeepLabv3+ model~\cite{chen2018encoder} to predict the successor graph from pose $\mathbf{q}$, which we parameterize as a heatmap $\mathbf{S}_\mathbf{q}$. To account for the additional \textit{Drivable} and \textit{Angles} input layers, which we feed into this model in addition to the RGB aerial image crop, we adapt the number of input layers of the model. We denote this model as SuccessorNet. To obtain per-pixel labeling of the successor graph in the image crop, we render the successor graph $\mathcal{S}_q$ as a heatmap in the crop by drawing along the graph edges with a certain stroke width. This heatmap highlights all regions in the aerial image that are reachable by an agent placed at pose $\mathbf{q}$. We train our SuccessorNet model with a binary cross-entropy loss. Finally, we skeletonize the predicted heatmap $\mathcal{\hat{S}}_q$ using a morphological skinning process~\cite{zhang1984fast} and convert the skeleton into a graph representation.

\subsection{Graph Exploration and Aggregation}
\label{subsec:graph-aggregation}

The approach described in the previous sections is capable of inferring the graph structure of the successor graph from a given query position. In this section, we illustrate how a complete lane graph can be obtained by running our AutoGraph model iteratively on its own predictions and by subsequently aggregating these predictions into a globally consistent graph representation. 

To this end, we leverage a depth-first exploration algorithm: We initialize our model by selecting start poses, which can either be chosen manually or obtained from our \textit{TrackletNet} model. We predict the successor graph from this initial position and repeatedly query our model along the successor graph. In the case of a straight road section, for each forward pass of our model, we add a single future query pose to the list of query poses to process. If a lane split is encountered, for each of the successor subgraphs starting at lane splits, we add a query pose to the list. If a lane ends or no successor graphs are found, the respective branch of the aggregated lane graph terminates and we query the next pose in the list. The exploration terminates once the list of future query poses is empty. In contrast to prior work~\cite{buchner2023learning}, where they aggregate the complete set of successor graphs according to an elaborate graph aggregation scheme, we instead only add graph nodes to the global graph where the virtual agent was placed at a given time. Therefore, we add edges between graph nodes according to the movement of the successor graph query position. This aggregation formulation simplifies the graph aggregation scheme since the number of nodes to integrate into the global graph is greatly reduced.

\begin{table*}
\centering
\footnotesize
\caption{Comparison of two baseline models with our AutoGraph approach for the Successor-LGP task. We evaluate on the test split of the UrbanLaneGraph dataset. Best model results are marked in \textbf{bold}.}
\label{tab:succ-lgp}
\begin{tabular}{p{3.1cm}|p{1cm}p{0.7cm}p{1.5cm}p{1.5cm}p{1.1cm}p{1.1cm}|c}
\toprule
Model & APLS $\uparrow$ &  IoU $\uparrow$ & TOPO P/R $\uparrow$ & GEO P/R $\uparrow$ & SDA$_{20}\uparrow$ & SDA$_{50}\uparrow$  & Human supervision \\
 \midrule
LaneGraphNet~\cite{zurn2021lane} & 0.179 & 0.063 & 0.0 / 0.0  &  0.0 / 0.0 &  0.0 & 0.0 & \checkmark \\
LaneGNN~\cite{buchner2023learning} & 0.202  & \textbf{0.347} &  \textbf{0.600} / \textbf{0.699} & \textbf{0.599} / \textbf{0.695} &  \textbf{0.227} & 0.377 &  \checkmark  \\
\hline
AutoGraph (ours) & \textbf{0.310} & 0.233 & 0.412 / 0.628 & 0.422 / 0.601 & 0.159 & \textbf{0.678} & \xmark    \\

\bottomrule
\end{tabular}
\end{table*}

\section{Dataset}
\label{sec:dataset}

We evaluate our proposed method on a large-scale dataset for lane graph estimation from traffic participants. We use the RGB aerial images and the ground-truth lane graph annotations from the UrbanLaneGraph dataset~\cite{buchner2023learning}. To obtain the traffic participant tracklets, we leverage the LiDAR dataset split of the Argoverse2~\cite{wilson2023argoverse} dataset. The dataset contains consecutive LiDAR scans for hundreds of driving scenarios. A single driving scenario entails approx. $\SI{20}{s}$ of real-world driving. We leverage the OpenPCDet~\cite{openpcdet2020} detection and tracking suite for LiDAR point clouds with a CenterPoint~\cite{yin2021center} model, pre-trained on the NuScenes dataset~\cite{caesar2020nuscenes}. We track the vehicle classes of \textit{Car}, \textit{Bus}, \textit{Trailer}, and \textit{Motorcycle}. Subsequently, we transform the respective LiDAR-centric tracklet coordinates to a global reference frame that is aligned with the aerial image coordinates. We smooth each tracklet with a mean filter approach to account for sensor noise and tracking inaccuracies. We call our tracklet dataset the \textit{UrbanTracklet} dataset and make it publicly available as an addition to the UrbanLaneGraph dataset~\cite{buchner2023learning}. In Tab.~\ref{tab:dataset-statistics}, we list all relevant metrics of our \textit{UrbanTracklet} dataset. In total, our dataset entails tracklets with an accumulated total length of approximately $\SI{12000}{km}$.

\begin{figure*}
\centering
\footnotesize
\setlength{\tabcolsep}{0.0cm}
    \begin{tabular}{m{1cm}m{16cm}}
\rotatebox{90}{Graph Annotation} &
\includegraphics[width=16cm]{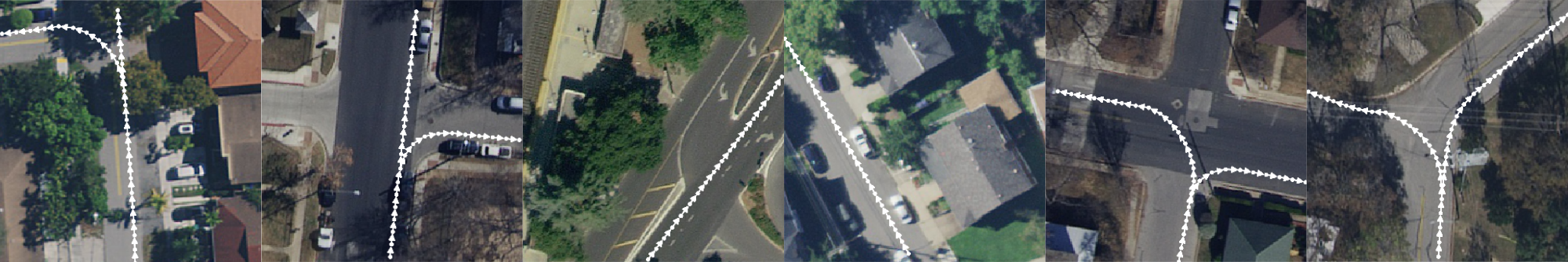} \\
\rotatebox{90}{AutoGraph-GT}    &
\includegraphics[width=16cm]{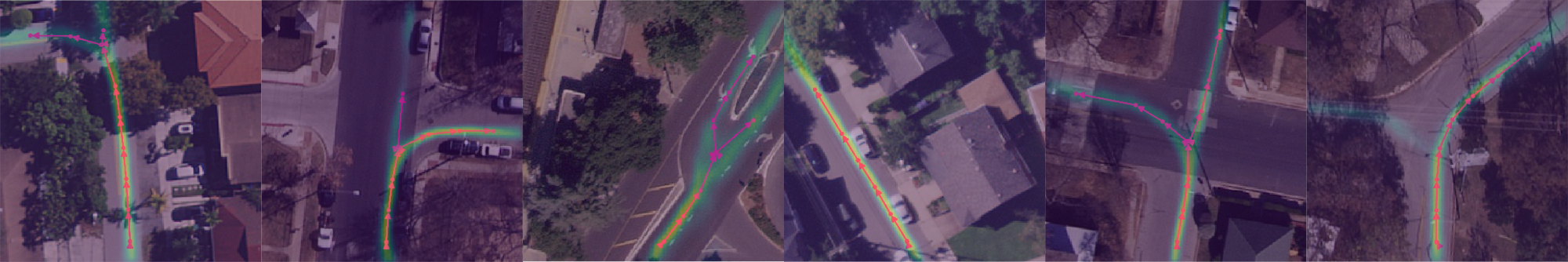} \\
\rotatebox{90}{AutoGraph}    &
\includegraphics[width=16cm]{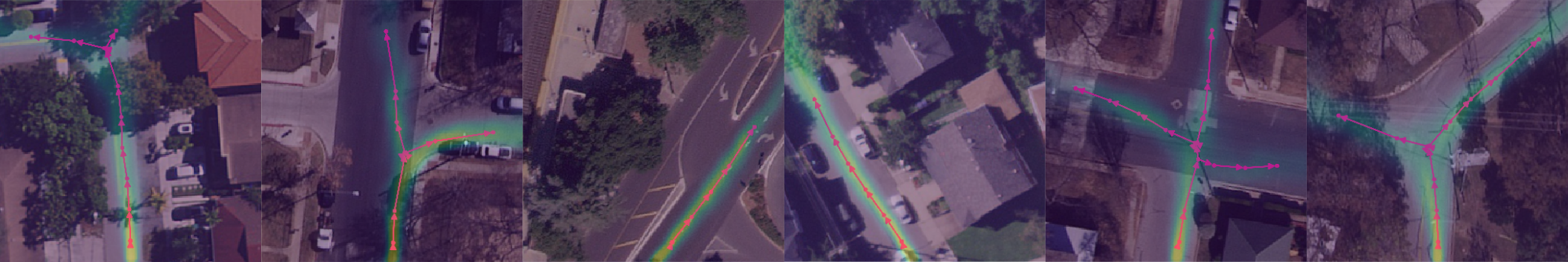} \\
    \end{tabular}
    \caption{Qualitative results of our models for the Successor-LGP task. We visualize the successor heatmap and the graph generated from it for our human-supervised model AutoGraph-GT and our tracklet-supervised model AutoGraph.}
    \label{fig:qualitative-successor} 
\end{figure*}

\begin{table}
\footnotesize
\caption{Key statistics of our \textit{UrbanTracklet} dataset}
\centering
 \begin{tabular}{p{2.1cm}|rr}
 \toprule
 \textbf{City} & \textbf{Number of tracklets}  & \textbf{Total tracklet length} \\
 \midrule
Austin, TX      & 287,306    & \SI{3,642}{km}   \\
Detroit, MI     &  73,232    & \SI{1,099}{km}   \\
Miami, FL       & 283,641    & \SI{3,312}{km}   \\
Palo Alto, CA   &  82,351    & \SI{1,050}{km}   \\
Pittsburgh, PA  &  34,505    & \SI{1,390}{km}   \\
Washington D.C. & 121,557    & \SI{1,469}{km}   \\
 \midrule
All             & 882,592    & \SI{11,962}{km}   \\
 \bottomrule
 \end{tabular}
\label{tab:dataset-statistics}
\end{table}

\section{Experimental Results}

\subsection{Implementation Details}

The TrackletNet and SuccessorNet have identical DeepLabv3+ architectures. The TrackletNet receives an RGB input image of shape $H \times W \times 3$  and outputs the \textit{Drivable} and \textit{Angles} layer map output. We use two separate decoders to produce the outputs. The drivable area segmentation has a resolution of $H \times W$ while the lane angle output has the size of $H \times W \times 2$. The training data used to train the two models is obtained from the dataset described in Sec.~\ref{sec:dataset}. We crop image segments of size $\SI{256}{px} \times \SI{256}{px}$ from the global aerial image. The crops are oriented along a randomly sampled tracklet at the bottom centre of each crop. To increase the efficacy of our aggregation method (see Sec.~\ref{subsec:graph-aggregation}), we require the successor graph prediction to be robust w.r.t.~perturbations in the position of the virtual agent. To provide more diverse samples with different positional variations, we randomly rotate the crop with an angle $\Delta \phi \sim \unif(-\pi/3, \pi/3)$.

Using this sampling method, a vast amount of samples can be generated since the aerial image can be cropped at arbitrary locations and orientations. For our experiments, we generate a total number of $\SI{1.5}{M}$ samples from all cities combined. We found that the lane graph complexity differs between different scenes, i.e., straight road sections have much simpler successor graphs than entries to roundabouts or multi-arm intersections. We found that a balanced mix between easy (successor graph has no splits) and hard  (successor graph has one or more splits) samples is beneficial.

\begin{table*}
\centering
\footnotesize
\caption{Ablation study with multiple model variants for the Successor-LGP task. We compare our AutoGraph model with our human-supervised model variant AutoGraph-GT and evaluate the influence of different map input layers on the model performance. We highlight the best-performing model with and without human supervision, respectively, in \textbf{bold}.}
\label{tab:succ-ablations}
\begin{tabular}{p{2.5cm}|p{1cm}p{0.7cm}p{1.1cm}p{1.1cm}|cp{1.1cm}p{1.5cm}p{1.2cm}}
\toprule
Model & APLS $\uparrow$ & IoU $\uparrow$ & SDA$_{20}\uparrow$ & SDA$_{50}\uparrow$  & Human supervision & Tracklet-Joining & Drivable map layer & Angles map layer \\
 \midrule
 AutoGraph-no-join & 0.344  & 0.232 & 0.052 & 0.498 & \xmark  & \checkmark & \xmark &  \xmark   \\
 AutoGraph & 0.310 & \textbf{0.233} & \textbf{0.159} & \textbf{0.678} & \xmark  & \xmark & \xmark &  \xmark   \\
AutoGraph+D & \textbf{0.349} & 0.230 & 0.063 & 0.510 & \xmark  & \checkmark & \checkmark &  \xmark   \\
AutoGraph+DA & 0.346 & 0.211 & 0.080 & 0.447 & \xmark  & \checkmark & \checkmark &  \checkmark   \\
\hline

AutoGraph-GT & 0.377 & \textbf{0.281} & 0.048 & 0.416 & \checkmark  & N/A & \xmark &  \xmark   \\
AutoGraph-GT+D & \textbf{0.418}  & 0.268  & \textbf{0.072} & 0.418  & \checkmark  & N/A & \checkmark &  \xmark   \\
AutoGraph-GT+DA & 0.409 & 0.269 &  0.059 & \textbf{0.463} & \checkmark  & N/A & \checkmark &  \checkmark   \\

\bottomrule
\end{tabular}

\end{table*}

\subsection{Tasks}
\label{sec:tasks}

Following Büchner~\etal~\cite{buchner2023learning}, we evaluate our approach on two complementary tasks: Successor Lane Graph Prediction (\textit{Successor-LGP}), and Full Lane Graph Prediction (\textit{Full-LGP}). In \textit{Successor-LGP}, we aim at predicting a feasible ego-reachable successor lane graph from the current pose of the virtual agent. In the task of \textit{Full-LGP}, we compare the complete lane graph in a local region to the ground-truth graph. We evaluate each task on the test images of the UrbanLaneGraph dataset~\cite{buchner2023learning}, which are not used for model training at any stage. For model evaluation, we use the metrics proposed by Büchner~\etal \cite{buchner2023learning}.

\subsection{Baselines}

To provide relevant comparisons and ablations demonstrating the efficacy of our AutoGraph approach, we compare it with a baseline model trained on ground-truth graph annotations, denoted as AutoGraph-GT. For this model, we use the ground-truth lane graph in places where we would otherwise query the recorded vehicle tracklets in our AutoGraph approach. This approach yields the ground truth lane graphs and successor lane graphs according to the graph annotations available in the dataset. The ground-truth lane graph annotations have none of the shortcomings of tracklet-based approaches, such as observation noise or erratic driving behaviour. We also compare to the previously proposed models LaneExtraction~\cite{he2022lane} and LaneGNN~\cite{buchner2023learning}. Note that other recent works by Colling \etal~\cite{colling2022hd} and Karlsson \etal~\cite{karlsson2023learning} are relevant works but do not aim at solving the task this work is concerned with and thus cannot be used for comparison.

\subsection{Task Evaluation}
\label{subsec:task-evaluation}

We evaluate our model on two tasks for lane graph estimation: Successor Lane Graph Prediction and Full Lane Graph Prediction.

\subsubsection{Successor Lane Graph Prediction}
\label{subsubsec:successor-lgp}

We evaluate the performance of our AutoGraph model and compare it with the recently proposed LaneGNN~\cite{zurn2021lane} and LaneGraphNet~\cite{buchner2023learning} models. Tab.~\ref{tab:succ-lgp} lists the model performances on the test split of the UrbanLaneGraph dataset. Our experiments indicate that the performance of our AutoGraph model is superior to the LaneGraphNet model in all metrics and is mostly on par with the recently proposed LaneGNN model. While it performs much better in the APLS and SDA$_{50}$ metric than the LaneGNN model, it is slightly inferior for the TOPO/GEO metrics and the Graph IoU metric. We hypothesize that the performance of our AutoGraph model could be further improved in scenes with road occlusions due to congested roads and overarching vegetation since our model struggles to predict accurate successor graphs in these regions. Specific treatment of such scenes in the model training schedule (i.e., active learning) might be beneficial.

Additionally, we perform ablation studies of multiple variants of our AutoGraph approach. The results are listed in Tab.~\ref{tab:succ-ablations}. In our AutoGraph-no-join variant, we do not join the tracklets (see Sec.~\ref{subsec:tracklet-parsing}), ignoring their proximity and their relative angles. Instead, we follow tracklets until they end or until they leave the image crop. We also do not use the \textit{Drivable} (D) and \textit{Angles} (A) model outputs but feed the aerial image directly into the SuccessorNet model. For our AutoGraph model variant, we use joined tracklets as per Sec.~\ref{subsec:tracklet-parsing} but omit the \textit{TrackletNet} auxiliary network. For our AutoGraph+D and AutoGraph+DA model variants, we add the \textit{Drivable} and \textit{Angles} model outputs, respectively. The model variant AutoGraph-GT does not use the tracklets of other traffic participants but is trained on ground truth human graph annotations, where we encode the successor graph as a heatmap instead of the raw graph representation as in the LaneGNN or LaneGraphNet models.

\begin{figure*}
    \centering
    \subfloat[Washington, D.C.]{\includegraphics[trim={0 0 0 1cm},clip,width=10cm]{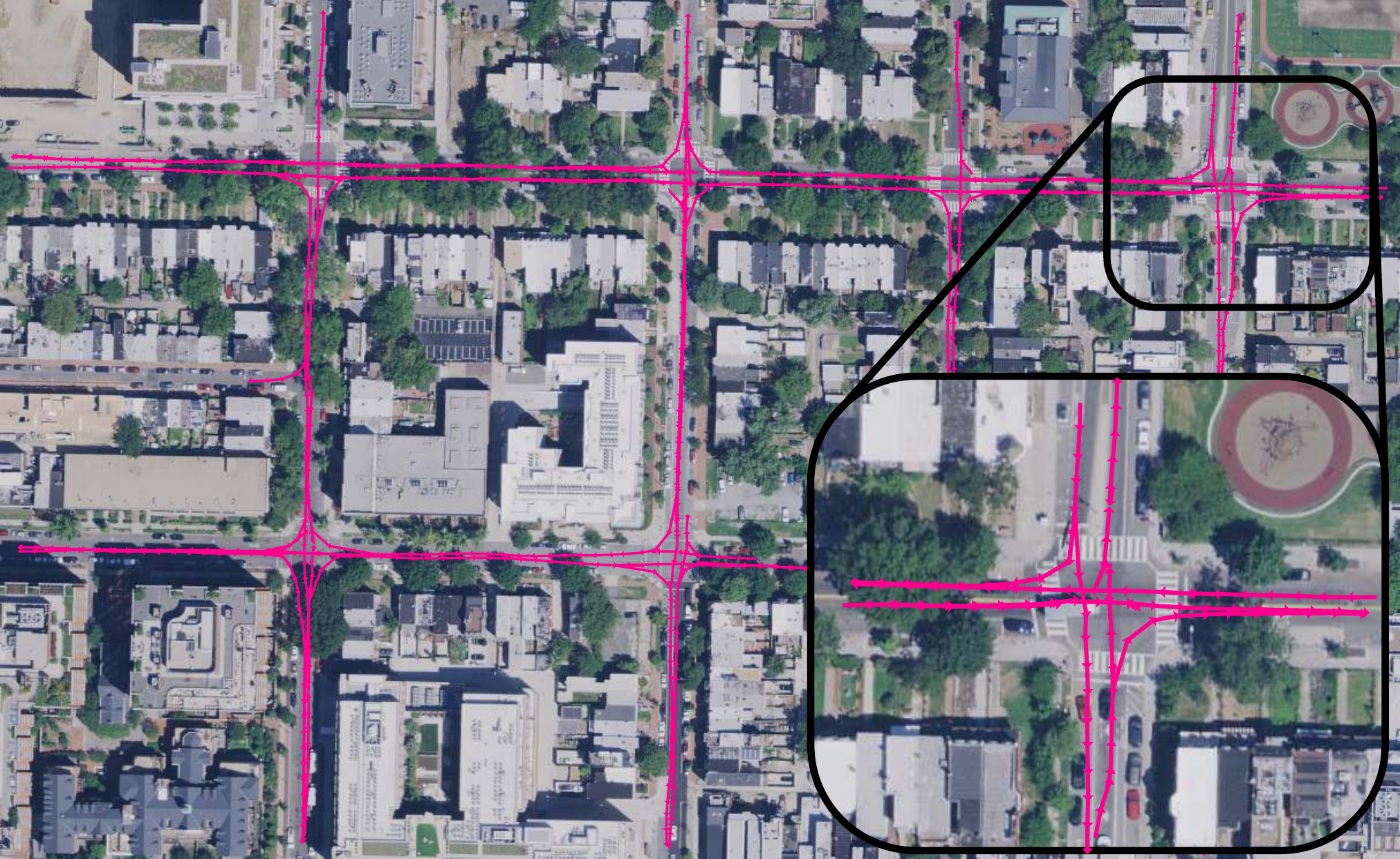}}%
    \subfloat[Miami, detail view]{\includegraphics[trim={0 0 0 0.7cm},clip,width=7.85cm]{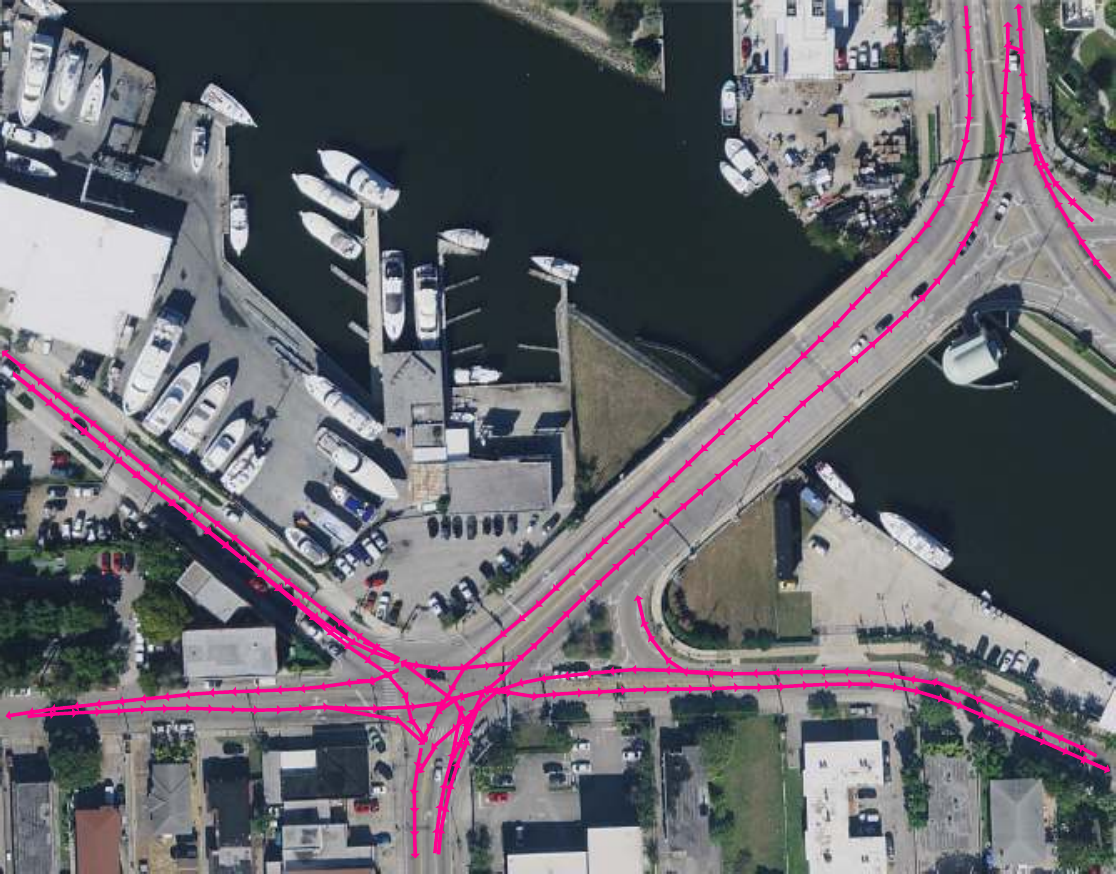}}%
\caption{Qualitative results on the Full-LGP task. We visualize predictions of our aggregated AutoGraph in pink color. Our aggregation scheme is capable of traversing challenging urban scenes featuring complex graph topologies with high accuracy.}
\label{fig:qualitative-full} 
\end{figure*}

Our ablation studies indicate that the AutoGraph-no-join method overall performs worse than our AutoGraph model variant. This indicates that joining tracklets to form more complete successor graphs helps produce higher-quality and more consistent annotations. Furthermore, the inclusion of the \textit{Drivable} map layer on top of the RGB layer improves model performance for some metrics. Adding the \textit{Angles} map layer in addition to the \textit{Drivable} layer does not consistently improve our evaluation metrics any further. Despite the additional information that is available about the scene if the \textit{Angles} map layer is included, the increased noise produced by imprecise angle estimates seems to outweigh the benefits of having additional information available. This result supports the results discussed in Zürn~\etal~\cite{zurn2021lane}, where additional input modalities did not significantly improve model performance. For our subsequent experiments, we opted to keep our \textit{Drivable} and \textit{Angles} model components despite the inconclusive results since drivable and angles maps are a useful model output for potential downstream tasks. Exemplarily, these outputs could also be aggregated and used to obtain a robust pixel-wise estimate for the prediction of drivable surface and lane orientation in large areas, similar to how we aggregate our successor graphs into a large graph structure.
 
For qualitative evaluation, we visualize predictions of our best-performing model in Fig.~\ref{fig:qualitative-successor} and the reference model AutoGraph-GT. We observe that both models are capable of modeling the multimodal spatial distribution of successor lanes efficiently. However, the AutoGraph-GT model shows more accurate heatmap outputs, since the annotations used for training stem from the ground-truth successor lane graph. 

To summarize, our experiments demonstrate that our AutoGraph model variants (trained on tracklets) perform overall similarly to our AutoGraph-GT model variants (trained on human lane graph annotations), indicating that vehicle tracklets recorded from a moving recording platform are suitable for training lane graph prediction models. In the APLS metric and the Graph IoU, the AutoGraph-GT model variant performs better than the AutoGraph model, presumably due to the higher annotation accuracy due to the better alignment of annotation with aerial images.

\subsubsection{Full Lane Graph Prediction}
\label{subsubsec:full-lgp}

For the Full Lane Graph Prediction task, we initialize our model on 10 initial poses per evaluation tile and run our aggregation scheme. We compare the performance of our best-performing model variant with the prediction results of LaneExtraction~\cite{he2022lane} and the aggregation module of LaneGNN~\cite{buchner2023learning}. Note that the number of initialization poses is much smaller compared to the number of initialization poses used for the LaneGNN model~\cite{buchner2023learning}. The results are listed in Tab.~\ref{tab:full-lgp}. We observe that for some metrics, our AutoGraph model achieves comparable or better performance than the human-supervised LaneGNN~\cite{buchner2023learning} or LaneGraphNet models~\cite{zurn2021lane}. Our model performs worse in the TOPO and GEO metrics. We note that our AutoGraph model struggles with road surface occlusions introduced by overarching vegetation. However, we emphasize that since our model uses fewer model initialization poses compared to LaneGNN~\cite{buchner2023learning}, a degradation in graph connectivity may be expected since lane graph regions in occluded areas may not be reached with an iterative aggregation scheme when no successor graph is found in a given frame.

Our qualitative evaluations show a high graph fidelity, recognizing most of the visible lanes and modelling their connectivity accurately. Fig~\ref{fig:qualitative-full} illustrates two exemplary lane graphs for the cities of Washington, D.C., and Miami. We observe that our approach is capable of accurately reconstructing the lane graph in visually challenging environments. Large scenes with multiple blocks are handled well and clearly reflect the underlying lane graph topology. The detailed view of a complex intersection in Miami illustrates that almost all major intersection arms are covered even in the presence of visual clutter such as water, boats, parking lots, and asphalt-colored roofs of buildings. Minor inaccuracies are produced at the five-armed intersection at the bottom of the aerial image, where not all connections between intersection arms are present in the inferred lane graph.

\begin{table}
\centering
\scriptsize
\caption{Evaluation on the Full-LGP task.}
\label{tab:full-lgp}
\begin{tabular}{p{2.2cm}|p{0.5cm}p{0.5cm}p{1.5cm}p{1.5cm}}
\toprule
Model & APLS &  IoU & TOPO P/R & GEO P/R \\
 \midrule
LaneExtraction~\cite{he2022lane} & 0.072 & 0.213 &  0.405 / 0.507  &   0.491 / 0.454  \\
LaneGNN~\cite{buchner2023learning} & 0.103 & \textbf{0.384} & 0.481 / \textbf{0.670}   &   \textbf{0.649} / \textbf{0.689}   \\
\hline
AutoGraph & \textbf{0.258} & 0.189  & \textbf{0.503} / 0.529 &  0.503 / 0.351 \\
\bottomrule
\end{tabular}
\end{table}

\section{Conclusion}
\label{sec:conclusion}

In this work, we presented a novel method for lane graph estimation in urban environments from traffic participant tracklets. We showed that our model, which is trained solely on data from tracked vehicles, is capable of predicting highly accurate lane graphs. We presented a tracklet processing scheme that allows us to use the observed tracklets of traffic participants as an annotation source to train our model. We demonstrated the efficacy of our approach on a large-scale lane graph dataset for which our approach demonstrated performance close to a ground-truth supervised baseline model. Future work will address adding pedestrians and bicycle tracklets to the approach for capturing more diverse annotations. Additionally, the improved handling of occluded roads appears to be a promising direction for future research.

\ifCLASSOPTIONcaptionsoff
  \newpage
\fi



%

{\small
\bibliographystyle{ieee_fullname}
\bibliography{root}
}

%








\end{document}